\pdfoutput=1
\documentclass[11pt]{article}

\usepackage[preprint]{acl}

\usepackage{comment}

\usepackage{times}
\usepackage{latexsym}

\usepackage[T1]{fontenc}
\usepackage[utf8]{inputenc}

\usepackage{microtype}

\usepackage{inconsolata}

\usepackage{graphicx}

\title{Improving Multi-Agent Debate with Sparse Communication Topology}

\author{
  \textbf{Yunxuan Li\textsuperscript{1$\dagger$}},
  \textbf{Yibing Du\textsuperscript{1}},
  \textbf{Jiageng Zhang\textsuperscript{1}},
  \textbf{Le Hou\textsuperscript{2}},
\\
  \textbf{Peter Grabowski\textsuperscript{1}},
  \textbf{Yeqing Li\textsuperscript{1}},
  \textbf{Eugene Ie\textsuperscript{1}}
\\
  \textsuperscript{1}Google
  \textsuperscript{2}Google DeepMind
}

\begin{document}
\maketitle
\renewcommand{\thefootnote}{\fnsymbol{footnote}}
\footnotetext[2]{Correspondence: yunxuanli@google.com}

\begin{abstract}
Multi-agent debate has proven effective in improving large language models quality for reasoning and factuality tasks. While various role-playing strategies in multi-agent debates have been explored, in terms of the communication among agents, existing approaches adopt a brute force algorithm -- each agent can communicate with all other agents. In this paper, we systematically investigate the effect of communication connectivity in multi-agent systems. Our experiments on GPT and Mistral models reveal that multi-agent debates leveraging sparse communication topology can achieve comparable or superior performance while significantly reducing computational costs. Furthermore, we extend the multi-agent debate framework to multimodal reasoning and alignment labeling tasks, showcasing its broad applicability and effectiveness. Our findings underscore the importance of communication connectivity on enhancing the efficiency and effectiveness of the ``society of minds'' approach.
\end{abstract}

\section{Introduction}
Large language models (LLMs) have demonstrated exceptional performance in natural language understanding and generation tasks. Recently a paradigm shift towards prompting LLMs has emerged as a significant and influential research area. By leveraging the in-context learning (ICL) capabilities of LLMs, these models can be adapted to various tasks such as reasoning, factuality, and AI feedback.

\begin{figure}[t]
  \includegraphics[width=\columnwidth]{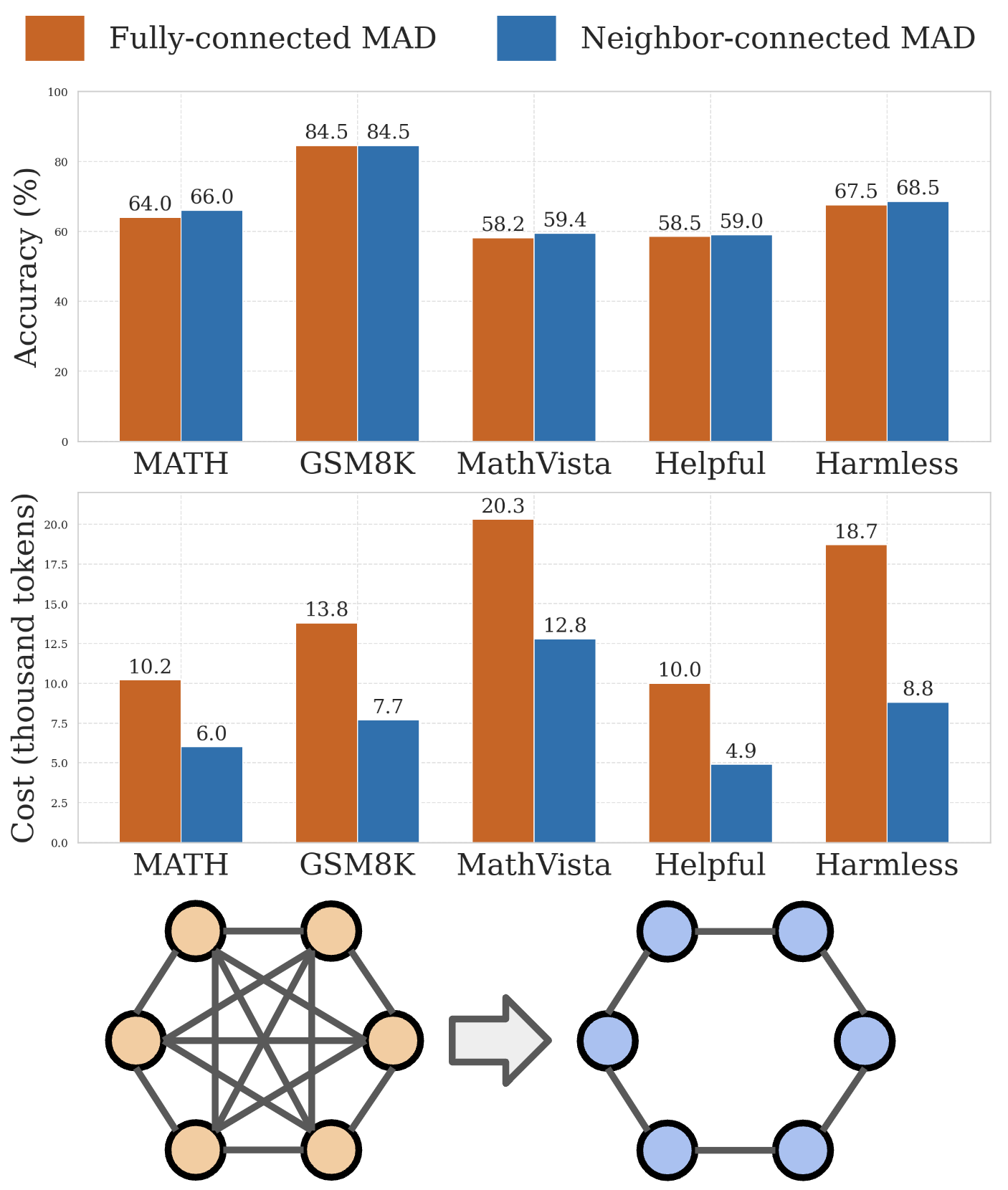}
  \caption{Accuracy (top) and inference input cost (middle) comparison of multi-agent debate system between fully-connected (bottom left) and neighbor-connected (bottom right) communication topologies.}
  \label{fig:fully_vs_neighbor}
\end{figure}

Several prompting methods have been developed to enhance LLM performance by optimizing their ICL capabilities. Notable techniques include Chain-of-Thought (CoT) \cite{wei2022chain}, self-consistency (SC) \cite{wang2022self}, and self-critique \cite{madaan2024self,welleck2022generating,shinn2024reflexion}. Recently, the multi-agent debate (MAD) framework is proven to be an innovative approach. Similar to a human discussion process, MAD employs multiple LLM agents to engage in discussions with one another, combining their reasoning and critical thinking abilities to produce high-quality results. Specifically, given a question, each agent first generates their own solutions to the question and then takes other agents' solutions as reference to update its own answer. This process can be repeated for several rounds. MAD has demonstrated significant improvement on factuality and reasoning tasks. While the debate process is highly productive, it is also very costly: As the number of LLM agents and debate rounds increase, the input context expands significantly.

Inspired by the intensive computational cost of MAD, a natural question arises: \textit{What if we reduce the number of reference solutions visible to each agent?} We conduct a systematic study on the sparsity of the multi-agent communication topology. Surprisingly, we find that sparse communication connectivity can deliver comparable or superior performance while significantly reducing inference costs. Figure \ref{fig:fully_vs_neighbor} presents a comparison between fully-connected MAD and neighbor-connected MAD. Compared to fully-connected MAD, neighbor-connected MAD achieves an improvement of $+2\%$ on the MATH dataset and maintains the same accuracy on GSM8K. Meanwhile, the average input token cost for reasoning tasks is reduced by over $40\%$.

MAD can also be a promising approach for Reinforcement Learning with AI Feedback (RLAIF) \cite{bai2022constitutional,lee2023rlaif} and weak-to-strong generalization \cite{burns2023weak}. By delivering better reward signals, MAD has the potential to significantly aid in aligning large language models. To assess this, we first extend the MAD framework to alignment labeling tasks, demonstrating its effectiveness compared to single-agent setups. Additionally, we verify that the advantages of sparsity observed in the reasoning tasks experiments also apply to alignment labeling tasks. Our experiments on the Anthropic-HH datasets
show an improvement of $+0.5\%$ in helpfulness and $+1.0\%$ in harmlessness, while reducing costs by $50.0\%$ and $53.3\%$, respectively.

We find that when agents are instantiated by different LLMs within the MAD framework, interactions between multiple LLMs result in weaker models being progressively strengthened through engagement with stronger models. In non-regular graph settings, assigning stronger LLMs to agents with higher centrality consistently yields better performance.

In summary, our contributions are listed as follows: (1) We demonstrate that sparse communication topology enhances both effectiveness and efficiency of the multi-agent debate framework; (2) We thoroughly evaluate sparse MAD for text-only and multimodal reasoning tasks, showing its advantage over standard MAD; (3) We extend the MAD framework to alignment labeling tasks, showing the effectiveness of standard MAD and further performance improvement introduced by sparse MAD; (4) We provide insights that explain the effectiveness of sparsity in MAD; (5) We find that assigning stronger LLMs to agents with higher centrality yields better overall performance in multiple LLM debate setup.

\section{Related Work}
\textbf{Multi-Agent Debate} MAD utilizes multiple LLM agents to discuss and debate with each other to generate and update the responses. It was first introduced by \citet{du2023improving}. Most of the MAD work focus on diversifying agents during the debate process. \citet{liang2023encouraging,park2023generative,li2023camel,chan2023chateval} highlight the importance of assigning different roles for agents. \citet{chen2023reconcile} diversifies agents' responses by instantiated with multiple LLMs. \citet{wang2024rethinking} proposes a method in which agents are divided into sub-groups and their discussion outcomes are later merged. Unlike other work, we aim to explore the effectiveness of sparse communication topology in MAD, and extend its applications to reasoning and alignment tasks.

\noindent \textbf{LLM Reasoning} Much work has been done to improve the reasoning ability of language models with prompting, including Chain-of-Thought \cite{wei2022chain} and its variants \cite{yao2024tree,besta2024graph}, problem decomposition \cite{zhou2022least}, reasoning ensemble \cite{wang2022self}, reasoner with verification \cite{cobbe2021training,wang2024multi,luo2023critique}.

\noindent \textbf{Multimodal Reasoning} With the recent advancements in vision-language models \cite{radford2021learning,yu2022coca, li2023blip,liu2024improved,lin2024moe}, multimodal large-language models (MLLMs) have demonstrated exceptional visual understanding capabilities. Several evaluation benchmarks have been proposed, such as VQAv2 \cite{balanced_vqa_v2}, OK-VQA \cite{okvqa}, ScienceQA \cite{lu2022learn}, MMMU \cite{yue2023mmmu}, and MathVista \cite{lu2023mathvista}. Similar to LLMs, MLLMs can also be improved through prompt-based methods. Various attempts have been made to enhance MLLMs in this manner \cite{zheng2024picture,ganz2024question,yang2023mm,zhao2024bba,zhang2023multimodal,chen2024large,Hu2024VisualSS}. Despite the effectiveness of these methods, they are often complex to design and implement. In this paper, we focus on improving multimodal reasoning using a multi-agent approach.

\noindent \textbf{AI Feedback} \citet{bai2022constitutional} first introduces the idea of RLAIF, in which LLM is used to annotate harmlessness preference. \citet{lee2023rlaif} compares various AI annotation methods. Recent work \citep{guo2024direct} also explores using AI feedback for online reinforcement learning, demonstrating the advantage of AI feedback for alignment research.

\section{Method}
\subsection{Communication Topology}
Communication topology of MAD refers to the connectivity structure among agents during the debate process. Communication topology can be represented as a graph $\mathcal{G}=(\mathcal{V}, \mathcal{E})$, where $\mathcal{V}$ is a set of agents and $\mathcal{E}$ is a set of communication channel. Presence of of any $(e_i, e_j)$ in $\mathcal{E}$ indicates that agent $i$ can access agent $j$'s previous round solutions during the debate process, and vice versa. We focus on static graphs in this work, while we also did exploratory experiments with dynamic graphs (Appendix \ref{appendix:probmad}).

We quantify the density of these graphs using the ratio of the number of edges to the maximum possible number of edges
$$D = \frac{2|\mathcal{E}|}{|\mathcal{V}| (|\mathcal{V}| - 1)}$$

\begin{figure}[t]
  \includegraphics[width=\columnwidth]{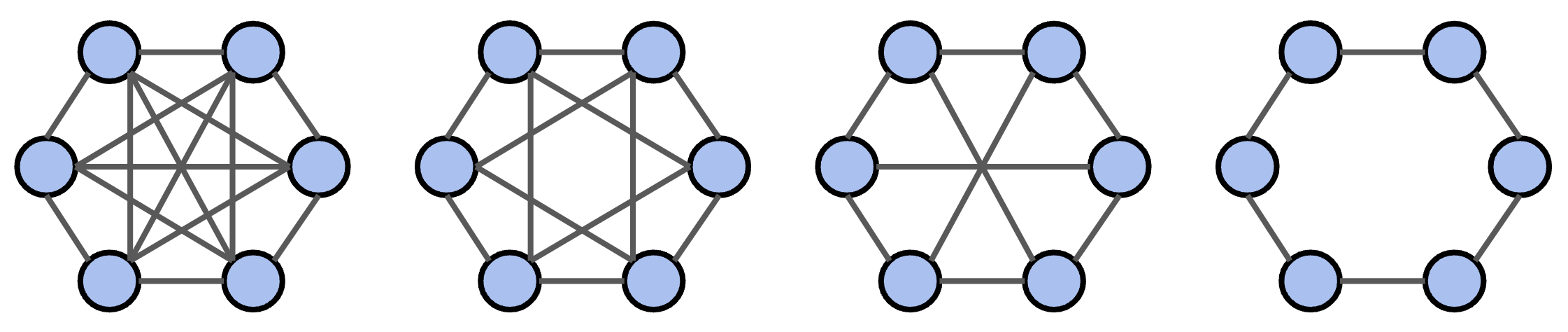} %{imgs/regular_graph}
  \caption{Communication topology of 6 agents with various sparsity. From left to right, the densities are 1 (fully-connected), $\frac{4}{5}$, $\frac{3}{5}$, and $\frac{2}{5}$ (neighbor-connected).}
  \label{fig:agent_6_sparsity_diagram}
\end{figure}

A lower value of $D$ indicates a sparser graph. In the standard MAD framework, agents are fully connected with each other, resulting in $D = 1$. In contrast, a neighbor-connected MAD has $|\mathcal{E}|=|\mathcal{V}|$, yielding $D=\frac{2}{|\mathcal{V}| - 1}$, which is a sparse graph. While the findings of this paper can be generalized to communication topology with an arbitrary number of agents, we focus on regular graphs where all agents have same degrees and are permutation invariant, with $|\mathcal{V}| = 6$ (Figure \ref{fig:agent_6_sparsity_diagram}). This choice is due to the limited spectrum of sparsity in scenarios with fewer agents and the significantly higher computational costs associated with analyzing scenarios with more agents. Additional experiment results with $|\mathcal{V}| = 4$ is shown in Appendix \ref{appendix:agent_4}.

\subsection{Multi-Agent Debate Process}
A typical MAD framework includes three steps:

\noindent \textbf{(1) Individual Response Generation}: In round 1, agents are initialized with LLMs, and then independently generate solutions to a given question. Typically a random decoding strategy is applied to diversify the solutions generated by agents.

\noindent \textbf{(2) Multi-agent Debate}: Starting round 2, each agent incorporates the responses of its connected peers from the previous round to critique or refine its own response. We utilize the standard \textit{Simultaneous-Talk} communication strategy \cite{chan2023chateval} to facilitate asynchronous computation. This debating process can occur over multiple rounds.

\noindent \textbf{(3) Reaching Consensus}: After the debate process, agents may still have differing solutions. In such cases, a majority vote is conducted among all agents to determine a consensus solution.

\section{Experiments Setup}
\subsection{Tasks}
%We aim to validate the effectiveness and efficiency of sparse MAD on reasoning and alignment labeling tasks. For reasoning tasks, we consider two text-only reasoning tasks: (1) MATH \cite{hendrycks2021measuring}: an arithmetic reasoning task containing challenging competition mathematics problems. We only use the \textit{algebra linear 1d composed} sub-task for simplicity. (2) GSM8K \cite{cobbe2021training}: a high quality grade school math reasoning task. For alignment labeling tasks, we consider Anthropic-HH dataset \cite{bai2022training}: human preference data on helpfulness and harmlessness.
We aim to validate the effectiveness and efficiency of sparse MAD on reasoning and alignment labeling tasks. For reasoning tasks, we consider two text-only reasoning tasks and one multimodal reasoning task: (1) MATH \cite{hendrycks2021measuring}: an arithmetic reasoning task containing challenging competition mathematics problems. We only use the \textit{algebra linear 1d composed} sub-task for simplicity. (2) GSM8K \cite{cobbe2021training}: a high quality grade school math reasoning task. (3) MathVista \cite{lu2023mathvista}: a benchmark designed to combine challenges from diverse mathematical and visual tasks. We only choose from \textit{free\_form} question type for consistency. For alignment labeling tasks, we consider Anthropic-HH dataset \cite{bai2022training}: human preference data on helpfulness and harmlessness.

\subsection{Models}
Our experiments utilize three publicly available models: GPT-3.5 \cite{gpt35}, GPT-4 \cite{gpt4}, and Mistral 7B \cite{jiang2023mistral}. Specifically, we employ GPT-3.5 for text-only reasoning tasks and GPT-4 for multimodal reasoning tasks. For alignment labeling tasks, we use both GPT-3.5 and Mistral 7B. We refrain from using GPT-4 for other tasks due to its significantly higher cost, which is approximately 10 times that of GPT-3.5. Additionally, we do not employ Mistral 7B for other tasks because of its inferior zero-shot performance on arithmetic reasoning. We randomly select 100 examples for each experiments involving GPT, and 500 examples for experiments with Mistral 7B.

\subsection{Baselines}
We compare sparse MAD against the following baselines:

(1) \noindent Chain-of-Thought (\textbf{CoT}): CoT prompting improves reasoning capabilities of LLMs with explicit intermediate reasoning steps.

(2) \noindent Self-consistency (\textbf{SC}): SC margins out intermediate reasoning paths by sampling diverse reasoning paths and selecting the most consistent answer.

(3) \noindent Existing MAD (\textbf{MAD} ($D=1$)): the standard approach for multi-agent debate, in which agents can communicate with all other agents with simultaneous-talk strategy. We also denote it as fully-connected MAD.

\subsection{Evaluation Metrics}
For reasoning tasks, we use the accuracy with respect to the ground truth answer to measure the quality of MAD. For alignment labeling tasks, we use \textit{AI Labeler Alignment} \cite{lee2023rlaif} to measure the accuracy of MAD labeling with respect to the human annotation.

Cost refers to the input inference cost of LLMs, which typically involves handling the autoregressive decoding mechanism and computational resources. Considering that advanced LLMs use a pay-per-token pricing model, we measure the inference cost by the number of input tokens.

\subsection{Variance Reduction}
Evaluating the significance of new communication topology compared to existing one typically involves running multiple random experiments to estimate the mean and variance of performance. However, this approach becomes impractical when the signal-to-noise ratio is low and each experimental run is computationally expensive. To address this, we employ two methods to reduce experimental variance and enhance the sensitivity of MAD with respect to the changes in communication topology: (1) As used by \citet{wang2024rethinking}, we reduce the temperature during language model decoding to stabilize performance. While we use the default temperature settings in API calls for most tasks, we lower the temperature to 0.25 for text arithmetic reasoning tasks to ensure robustness. (2) We employ conditional variance reduction \cite{ross2002simulation}. Observing that most of the variance arises from the first round of individual responses, we first generate a set of initial agent responses and then fix them in all subsequent debate processes across various communication topology designs. This approach effectively minimizes variance and improves the reliability of our experimental results.

\section{Experiments: MAD with Single LLM}
\subsection{MAD on Text Reasoning Tasks}
We build on existing work on MAD, exemplified by reasoning tasks, by showing the advantages of sparse MAD on top of the proven advantage of fully-connected MAD. Sparse MAD significantly saves computational cost while preserving comparable or better performance.

\textbf{Sparse MAD has similar or higher accuracy with significant cost saving on reasoning tasks}: For both the MATH and GSM8K tasks, we demonstrate that sparse MAD produces comparable or better accuracy than fully-connected MAD, while significantly cutting down inference costs. Both fully-connected and sparse MAD setups outperform Chain-of-Thought and self-consistency methods. Specifically, in the MATH task, fully-connected MAD shows a $+4.0\%$ quality gain over self-consistency, while sparse MAD configurations achieve accuracy improvements ranging from $+3.0\%$ to $+7.5\%$ (Table \ref{table:math}). Similarly, in the GSM8K task, fully-connected MAD demonstrates a $+4.5\%$ quality gain over self-consistency, whereas sparse MAD achieves accuracy improvements between $+3.5\%$ and $+6.5\%$ (Table \ref{table:gsm8k}). Furthermore, sparse MAD setups reduce costs by up to $-41.5\%$ and $-43.5\%$, respectively. It is important to note that we exclusively use the GPT-3.5 model because Mistral 7B performs poorly on these challenging tasks in a zero-shot setting.

\begin{table}
  \centering
  \begin{tabular}{l|cc}
    \hline
    \textbf{Method} & Accuracy & Cost Saving \\
    \hline
    CoT & 58.0 $\pm$ 2.0 & - \\
    SC & 60.0 & - \\
    \hline
    MAD ($D=1$) & 64.0 $\pm$ 1.4 & baseline  \\
    MAD ($D=4/5$) & \textbf{67.5 $\pm$ 2.0} & $-$14.6\% \\
    MAD ($D=3/5$) & 63.0 $\pm$ 1.8 & $-$29.2\% \\
    MAD ($D=2/5$) & 66.0 $\pm$ 2.3 & $-$41.5\% \\
    \hline
  \end{tabular}
\caption{Comparison of accuracy and cost savings of MAD against baseline methods on the MATH dataset. All experiments were conducted using the GPT-3.5 model.}
\label{table:math}
\end{table}

\begin{table}
  \centering
  \begin{tabular}{l|cc}
    \hline
     % & \multicolumn{2}{c}{\textbf{GPT-3.5}} \\
    \textbf{Method} & Accuracy & Cost Saving \\
    \hline
    CoT & 77.5 $\pm$ 4.2 &  - \\
    SC  & 80.0 &  - \\
    \hline
    MAD ($D=1$) & 84.5 $\pm$ 1.5 & baseline \\
    MAD ($D=4/5$) & 83.5 $\pm$ 0.5 & $-$12.7\% \\
    MAD ($D=3/5$) & \textbf{86.5 $\pm$ 1.5} & $-$29.1\% \\
    MAD ($D=2/5$) & 84.5 $\pm$ 0.8 & $-$43.6\% \\
    \hline
  \end{tabular}
 \caption{Comparison of accuracy and cost savings of MAD against baseline methods on the GSM8K dataset. All experiments were conducted using the GPT-3.5 model.}
\label{table:gsm8k}
\end{table}

\subsection{MAD on Multimodal Reasoning Task}
MAD on multimodal reasoning tasks also demonstrates notable improvements compared to Chain-of-Thought and self-consistency approaches. This suggests that MLLMs like GPT-4o can effectively integrate step-by-step reasoning with visual content to enhance final answers. Similar to text reasoning experiments, we examine various sparse MAD configurations and report their performance.

\textbf{Sparse MAD retains performance while introducing significant cost savings on multimodal reasoning tasks.} For the MathVista task, we evaluate different MAD configurations, comparing them to each other as well as to Chain-of-Thought (CoT) and self-consistency methods (Table \ref{table:mathvista}). We find that sparse MAD achieves similar or slightly better accuracy compared to fully-connected MAD, with both outperforming CoT and self-consistency. The best sparse MAD configuration achieves a $+$1.2\% improvement over fully-connected MAD and a $+$6.4\% improvement over self-consistency. Additionally, sparse MAD provides substantial cost savings, reducing the total number of tokens used by up to 33.1\%. Given that multimodal inputs are typically much larger than textual inputs (e.g., in GPT-4o, each image costs at least 225 tokens and can grow to 400$+$, 600$+$, or more tokens), we observe a total reduction of 40.6\% in token usage, excluding the input image tokens.

\begin{table}
  \centering
  \begin{tabular}{l|cc}
    \hline
     % & \multicolumn{2}{c}{\textbf{GPT-3.5}} \\
    \textbf{Method} & Accuracy & Cost Saving \\
    \hline
    CoT & 52.4 $\pm$ 2.6 &  - \\
    SC  & 53.0 &  - \\
    \hline
    MAD ($D=1$) & 58.2 $\pm$ 1.5 & baseline \\
    MAD ($D=4/5$) & 57.8 $\pm$ 1.9 & $-$9.1\% ($-$11.5\%) \\
    MAD ($D=3/5$) & 55.4 $\pm$ 0.9 & $-$20.0\% ($-$24.7\%) \\
    MAD ($D=2/5$) & \textbf{59.4 $\pm$ 0.6} & $-$33.1\% ($-$40.6\%) \\
    \hline
  \end{tabular}
 \caption{Comparison of accuracy and cost savings of MAD against baseline methods on the MathVista dataset. All experiments were conducted using the GPT-4o model with the default temperature $T=1$. The cost saving percentages in parenthesis are computed without multimodal inputs.}
\label{table:mathvista}
\end{table}

\subsection{MAD on Alignment Labeling Tasks}
Alignment labeling tasks involve annotating preferences between pairs of responses generated for a given question. Our prompt consists of three parts: (1) a system prompt that informs the LLM of its role as a rater and specifies the required answer formatting; (2) a question description providing the context of the question; and (3) an ending instruction that constrains the answer length and reiterates the formatting requirements. During the debate, reference solutions are included before the ending instruction. See \ref{appendix:alignment_prompt_template} for more details.

We use \textit{AI Labeler Alignment} \cite{lee2023rlaif} to measure the accuracy of MAD labeling with respect to the human annotation. To prevent potential position bias, we randomly assign the chosen response to either the (A) or (B) option. We report the accuracy and inference cost of MAD with various level of sparsity in Table \ref{table:helpfulness} for helpfulness and Table \ref{table:harmlessness} for harmlessness.

\begin{table*}
  \centering
  \begin{tabular}{l|cc|cc}
    \hline
     \textbf{Method} & \multicolumn{2}{c|}{\textbf{GPT-3.5}} & \multicolumn{2}{c}{\textbf{Mistral 7B}} \\
     & Accuracy & Cost Saving & Accuracy & Cost Saving \\
    \hline
    CoT & 56.5 $\pm$ 3.1 & - & 60.8 $\pm$ 1.2  & - \\
    Self-Consistency  & 57.0 & - & 62.6 & - \\
    \hline
    MAD ($D=1$) & 58.5 $\pm$ 1.7 & baseline & 65.5 $\pm$ 0.6 & baseline \\
    MAD ($D=4/5$) & \textbf{59.0 $\pm$ 1.8} & $-$17.5\% & 65.6 $\pm$ 0.9 & $-$18.3\% \\
    MAD ($D=3/5$) & 57.0 $\pm$ 1.6 & $-$32.5\% & 64.6 $\pm$ 0.6 & $-$35.2\% \\
    MAD ($D=2/5$) & \textbf{59.0 $\pm$ 1.4} & $-$50.0\% & \textbf{66.6 $\pm$ 0.5} & $-$53.5\% \\
    \hline
  \end{tabular}
\caption{AI labeler alignment accuracy and cost savings of MAD compared with baselines on the helpfulness dataset for GPT-3.5 and Mistral 7B models.}
\label{table:helpfulness}
\end{table*}

\begin{table*}
  \centering
  \begin{tabular}{l|cc|cc}
    \hline
     \textbf{Method} & \multicolumn{2}{c|}{\textbf{GPT-3.5}} & \multicolumn{2}{c}{\textbf{Mistral 7B}} \\
     & Accuracy & Cost Saving & Accuracy & Cost Saving \\
    \hline
    CoT & 66.0 $\pm$ 4.8 & - & 58.2 $\pm$ 2.0 & - \\
    Self-Consistency  & 67.0 & - & 60.0 & - \\
    \hline
    MAD ($D=1$)    & 67.5 $\pm$ 0.6 & baseline & 60.7 $\pm$ 0.3 & baseline \\
    MAD ($D=4/5$)  & 67.0 $\pm$ 0.8 &  $-$17.3\% & \textbf{62.2 $\pm$ 0.2} & $-$17.9\% \\
    MAD ($D=3/5$)  & 67.5 $\pm$ 1.0 & $-$34.7\% & 60.4 $\pm$ 0.4 & $-$34.3\% \\
    MAD ($D=2/5$)  & \textbf{68.5 $\pm$ 0.7} & $-$53.3\% & 61.7 $\pm$ 0.2  & $-$52.2\% \\
    \hline
  \end{tabular}
\caption{AI labeler alignment accuracy and cost savings of MAD compared with baselines on the harmlessness dataset for GPT-3.5 and Mistral 7B models.}
\label{table:harmlessness}
\end{table*}

\textbf{MAD outperforms single-agent on alignment labeling tasks}: We find that MAD consistently outperforms single-agent methods, including CoT and self-consistency. On the helpfulness task, fully-connected MAD achieves a $+1.5\%$ and $+2.9\%$ improvement over self-consistency for GPT-3.5 and Mistral 7B models, respectively. On the harmlessness task, fully-connected MAD achieves a $+0.5\%$ and $+0.7\%$ improvement over self-consistency for GPT-3.5 and Mistral 7B models, respectively. These results suggest that the additional debate process in MAD, followed by majority voting, allows agents to incorporate perspectives from others and refine their opinions toward the correct answers during the debate process.

\textbf{Sparse MAD can perform better with lower inference costs}: Most sparse MAD configurations perform as well as or better than the fully-connected MAD, with at least one sparse topology outperforming the fully-connected MAD. Depending on the task, sparse MAD with GPT-3.5 can enhance performance by approximately $+0.5\%$ to $+1.0\%$, and sparse MAD with Mistral 7B can improve performance by about $+1.1\%$ to $+1.5\%$. Additionally, sparse MAD can reduce costs by up to $-53.3\%$ and $-53.5\%$, respectively.

We observed that GPT-3.5 exhibits lower alignment accuracy compared to Mistral 7B on the helpfulness task. We attribute this discrepancy to the differences in pre-training and post-training corpora between the two models, which may lead to varying default preferences in a zero-shot setting. While we hypothesize that few-shot prompting techniques could mitigate this issue, exploring this is beyond the scope of this work.

\subsection{Why Does Sparse MAD Work?}
The common explanation for the effectiveness of MAD against single-agent setups is that agents can consider different perspectives before arriving at an answer. However, our experiment on the effectiveness of sparse MAD seems challenge this intuition. In this section, we aim to explain why sparse MAD can achieve comparable or even superior performance.

\begin{figure}[t]
  \includegraphics[width=\columnwidth]{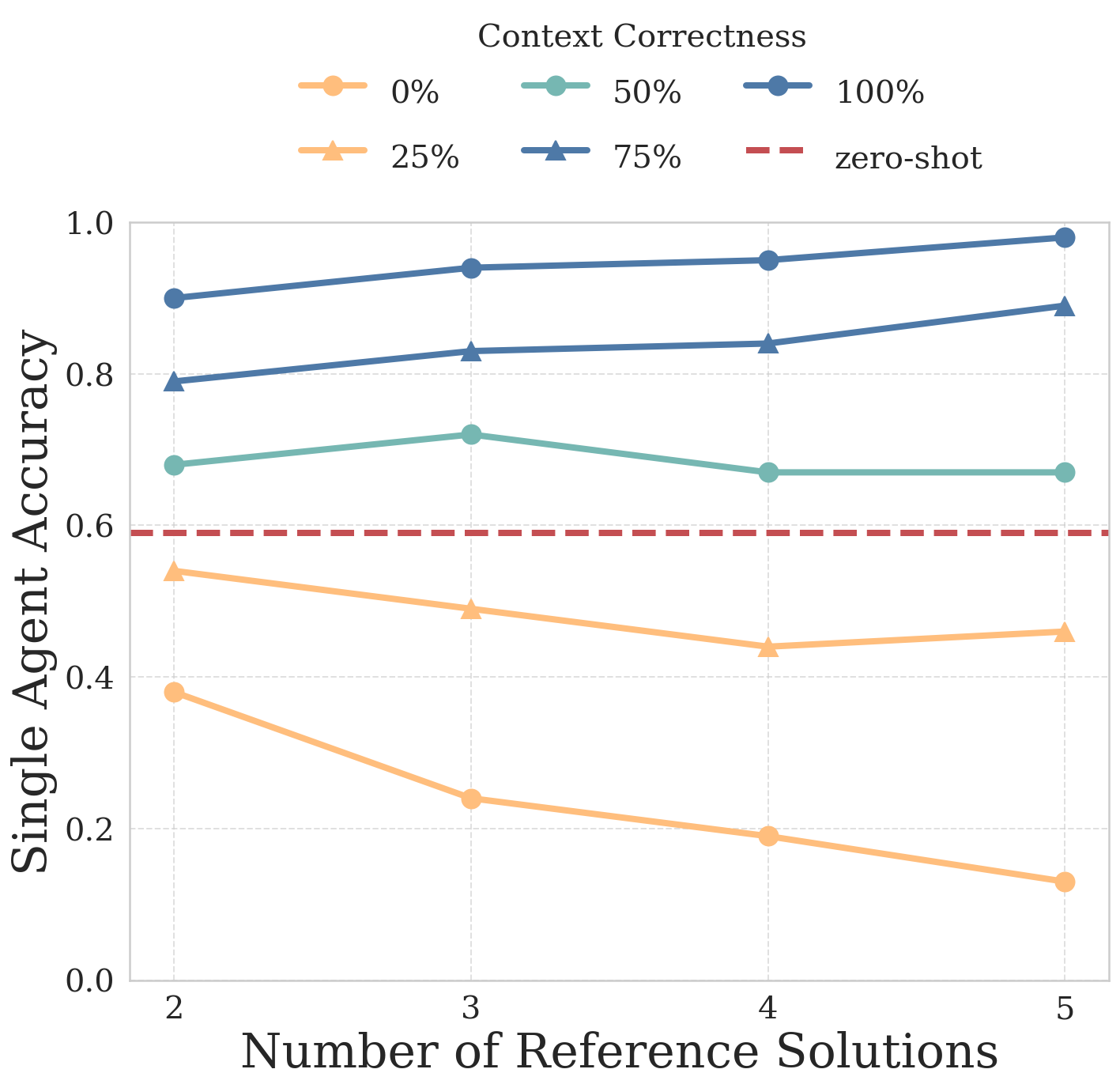}
  \caption{Probability of a single agent generating correct answers given $n$ reference solutions, with $p$ representing the correctness of these solutions. Monte Carlo sampling was performed on three questions, each with 100 runs.}
  \label{fig:analysis_trend}
\end{figure}

\begin{figure}[t]
  \includegraphics[width=\columnwidth]{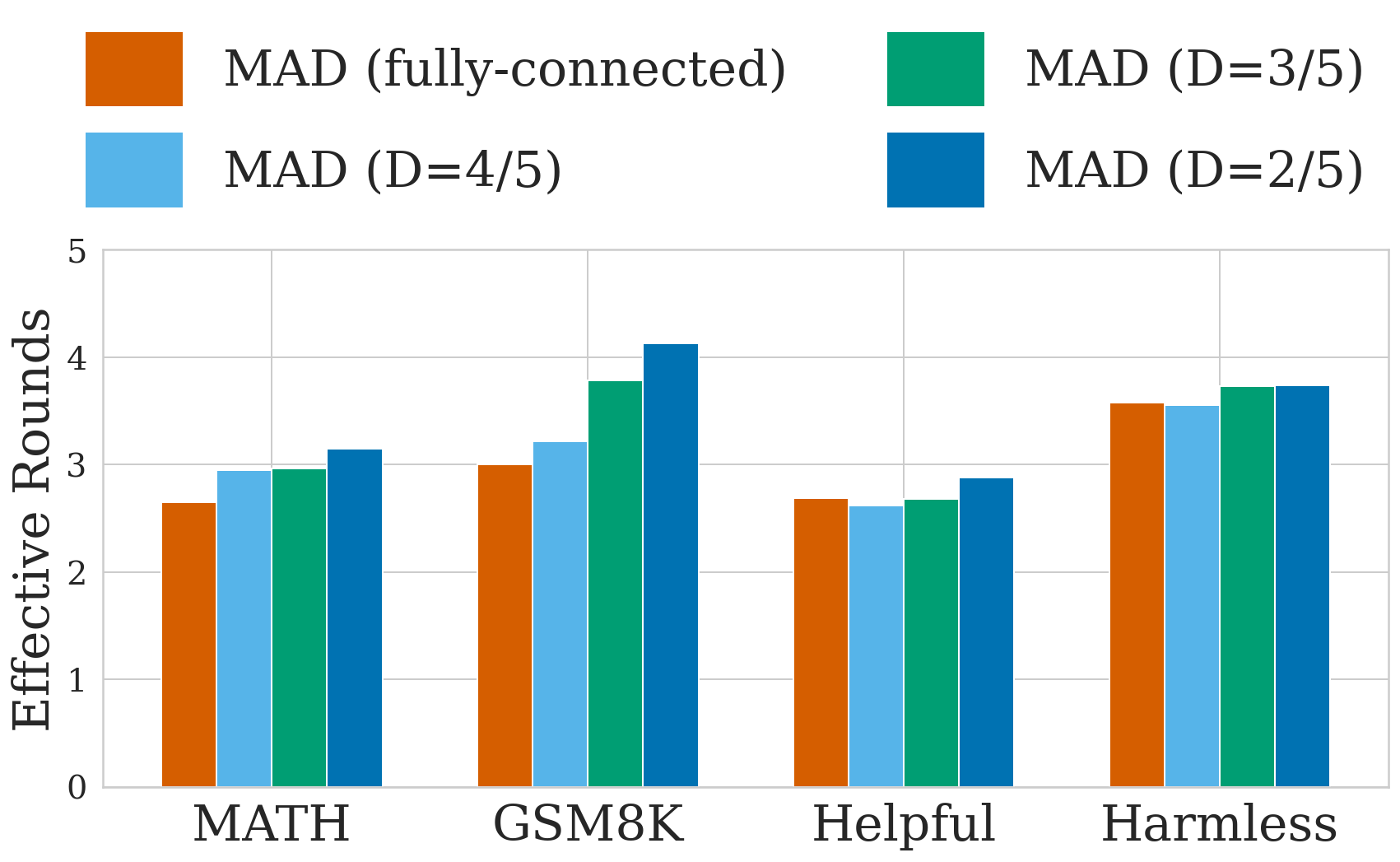}
  \caption{Effective debate rounds for each topology design in reasoning and alignment labeling tasks.}
  \label{fig:analysis_effective_debate}
\end{figure}

\textbf{Impact of incorrect reference solutions}: In a MAD framework, we define $Q(n, p)$ as the probability that a single agent delivers correct answers given $n$ reference solutions, where $p$ percentage of these are correct. This probability, $Q(n, p)$, can be estimated using Monte Carlo sampling with constructed in-context reference solutions. As a case study, we selected three questions from the GSM8K dataset and estimated $Q(n, p)$ for $n \in \{2,3,4,5\}$ and $p \in \{0\%, 25\%, 50\%, 75\%, 100\%\}$. Here, the choice of $n$ corresponds to the single-agent scenarios in MAD with $D=\frac{2}{5}, \frac{3}{5}, \frac{4}{5}, 1$. Results shown in Figure \ref{fig:analysis_trend} indicate that for easier questions, where most reference solutions are correct, an increase in the number of observed reference solutions (namely MAD becomes denser) improves the likelihood of the agent arriving at the correct answer. Conversely, for more difficult questions, where most agents do not provide correct answers, an increase in the number of observed reference solutions tends to mislead the agent into choosing incorrect answers, thereby drastically reducing the likelihood of reaching a correct response.

\textbf{Sparser MAD allows more rounds of effective debate}: We observe that once all agents converge on the same answer, it becomes highly unlikely for any of them to change their decision. We define the number of effective debates as the number of rounds before all agents reach the same answer. Figure \ref{fig:analysis_effective_debate} illustrates the effective number of debate rounds for various topologies in reasoning and alignment labeling tasks. Our results show that sparse MAD tends to sustain longer debates before achieving consensus, indicating that sparse MAD allows for more extensive deliberation and in-depth discussion. We observe there are similar findings in the Chain-of-Thought prompting \cite{jin2024impact} and MAD \cite{du2023improving} that the increase of reasoning length can significantly improve the performance.

\section{Experiments: MAD with Multiple LLMs}
Previous sections focus on the MAD with agents instantiated by the same LLM. In this section, we explore the scenario when multiple LLMs are available. With agents instantiated by different LLMs, the permutation invariance symmetry is broken, and the regular graph may not be optimal. A natural question is: \textit{how to design the communication topology given a MAD framework of $N$ agents, in which $M$ instantiated by the stronger LLM and $N - M$ instantiated by the weaker LLM?}

\begin{figure}[t]
  \includegraphics[width=\columnwidth]{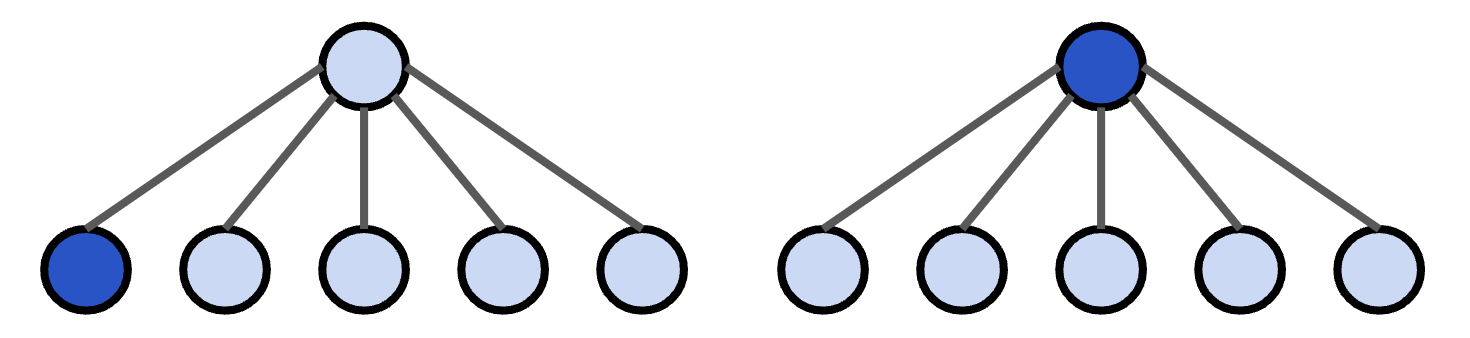}
  \caption{Isotropic communication topology with two setups: the stronger LLM has low centrality (left) and high centrality (right).}
  \label{fig:A5}
\end{figure}

\begin{table}
  \centering
  \begin{tabular}{l|cc}
    \hline
     \textbf{Centrality}  & \multicolumn{2}{c}{\textbf{Accuracy}}  \\
      & SC & Isotropic MAD \\
    \hline
    High & 64.0 & \textbf{67.0 $\pm$ 0.8} \\
    Low  & 64.0 & 65.8 $\pm$ 0.5 \\
    \hline
  \end{tabular}
\caption{Comparison of accuracy depending on where a stronger LLM is placed in debate, using the Harmlessness task as example. In both cases, there are 5 Mistral models and 1 GPT-3.5 Model. Accuracy of Isotropic MAD is calculated as the average over debate rounds.}
\label{table:multiple_llm_result}
\end{table}

\begin{figure}[t]
  \includegraphics[width=\columnwidth]{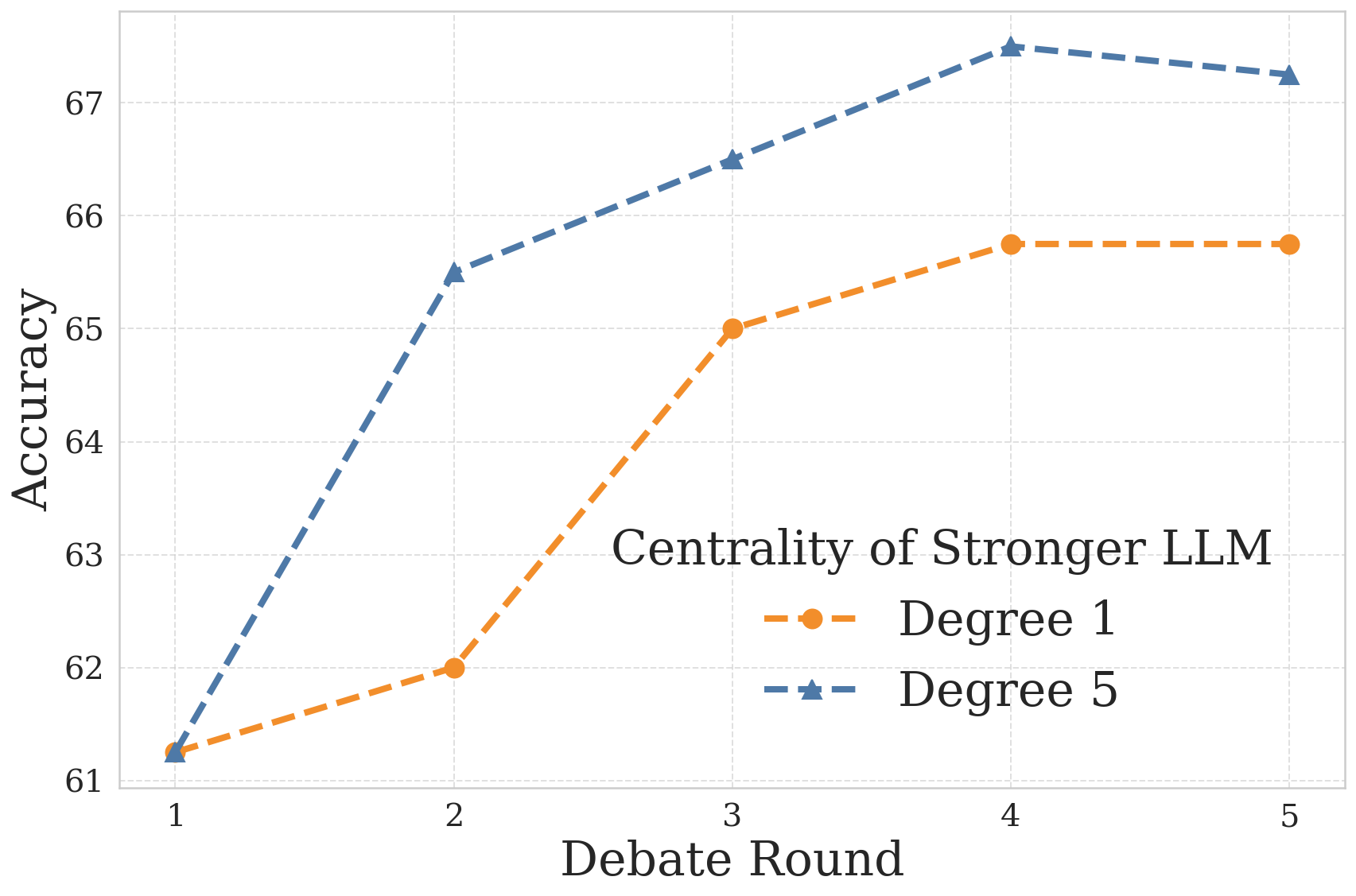}
  \caption{Average accuracy of weaker agents across different debate rounds.}
  \label{fig:multiple_llm_trend}
\end{figure}

\textbf{Assigning stronger LLMs to agents with higher centrality yields better performance}: We conducted experiments on harmlessness alignment labeling task, involving 6 agents, with 1 agent utilizing GPT-3.5 (the stronger LLM) and the remaining 5 agents utilizing Mistral 7B (the weaker LLM). We tested two setups on the isotropic communication topology: one where the stronger LLM had a degree of 1 (indicating low centrality) and another where it had a degree of 5 (indicating high centrality), as illustrated in Figure \ref{fig:A5}. The experimental results presented in Table \ref{table:multiple_llm_result} show that positioning the stronger LLM at a node with higher centrality (degree of 5) leads to better performance ($+3.0\%$ improvement) compared to placing it at a node with lower centrality (degree of 1) which resulted in a $+1.8\%$ improvement.

The results above underscore the importance of information flow in the design of communication topology. Figure \ref{fig:multiple_llm_trend} illustrates the average accuracy of weaker agents with respect to the number of debate rounds. When the stronger agent has a degree of 5, it can effectively disseminate its knowledge to weaker agents in just one debate round, resulting in a sharp increase in the average accuracy of weaker LLMs. In contrast, when the stronger agent has a degree of 1, the process requires two rounds: first, the information is transmitted to the central weaker agent in the first debate round (round 2), which then shares it with other weaker agents in the next round (round 3). This two-step process leads to greater information loss.

\section{Conclusion}
In this paper, we show that sparse communication topologies can improve the multi-agent debate framework significantly. Our results indicate that sparse MAD configurations achieve comparable or superior performance to standard MADs while greatly reducing computational costs. We also extend the MAD framework to alignment labeling tasks, demonstrating the benefits of MADs over single-agent setups and self-consistency and further highlighting the benefits of sparse MADs over fully-connected configurations. We present case-study insights that explain the effectiveness of sparse MADs. Additionally, we investigate the impact of communication topology design with multiple large language models (LLMs), finding that assigning stronger LLMs to more connected agents enhances overall performance.

In summary, our work paves the way for more efficient and effective multi-agent systems by leveraging sparse communication topologies. Future studies could focus on deepening our understanding of the underlying mechanisms and developing strategies for optimal topology design in multi-agent frameworks.

\section{Ethical Considerations}
In this work, several ethical considerations were addressed to ensure the integrity and responsible use of the system:

\noindent \textbf{Public Datasets}:  The framework was built using publicly available datasets that are designed for academic research. We strictly adhered to ethical guidelines by not using any personal or confidential data.

\noindent \textbf{License}: Only public APIs that offer appropriate licensing were utilized. This ensures that all external tools are used in a lawful and ethical manner.

\noindent \textbf{AI assistant}: AI tools were employed solely for polishing writing and correcting grammar. The AI was not used to generate content or ideas, maintaining the authenticity and originality of the research work.

\section{Limitations}
While our study provides valuable insights into the communication topology analysis of multi-agent debate, several limitations must be acknowledged:

Our analysis is primarily based on static graphs where the communication topology remains unchanged throughout the debate rounds. This constraint simplifies the analysis, but ignores the dynamic nature of real-world communication networks. Additionally, our study focuses on prompt design under a zero-shot setting, utilizing only publicly available GPT and Mistral models. This narrow scope may not fully capture the variability and adaptability present in more diverse agent populations. Furthermore, we confined our analysis to regular graphs, which do not encompass the full spectrum of potential graph configurations. Future work should consider dynamic graphs, a broader range of models, and varied graph connectivity to better reflect the evolving and complex nature of multi-agent interactions.

Our study relies on a subset of academic datasets due to limited data access as well as computational constraints. While these datasets provide a valuable foundation for analyzing communication graph dynamics in multi-agent debates, they may not fully represent the diversity and complexity found in broader real-world data. The restricted scope limits our ability to generalize findings across different domains and contexts. Future research should aim to include a wider range of datasets, potentially leveraging more efficient computational resources, to enhance the robustness and applicability of our findings.

We lack a rigorous theoretical proof explaining why sparse connectivity can lead to better performance. This gap in our understanding limits our ability to generalize our findings and apply them with confidence in various settings. Secondly, we do not have a definitive method for determining the optimal topology design, which is crucial for maximizing the efficiency and effectiveness of multi-agent systems. Addressing these questions is essential for future research. Potential explanations might involve theoretical insights, social and psychological dynamics, or a combination of these factors. Additionally, fine-tuning models could offer further clarity and aid in optimizing communication topology. Future work should aim to develop robust theoretical frameworks and empirical strategies to better understand and leverage communication topology in multi-agent debates.

The multi-agent debate framework holds significant potential for various real-world applications. However, it also carries the risk of misuse, including the dissemination of biased information or misinformation. Additionally, the framework requires substantial computational resources, which could impact energy consumption and environmental sustainability. Future research should focus on developing robust, trustworthy, and energy-efficient multi-agent systems to mitigate these risks and ensure ethical, reliable, and sustainable outcomes.

% \bibliography{anthology,main,custom}
\bibliography{main}

%\newpage
\appendix

\section{Prompt Templates}
\label{appendix:alignment_prompt_template}

\subsection{Text Reasoning Tasks}

\noindent \textbf{System Prompt}:

\noindent You are a helpful assistant with expertise in mathematics and reasoning. Your task is to assist in solving a math reasoning problem by providing a clear and detailed solution. Limit your output within 100 words, and your final answer should be a single numerical number, in the form of \{\{\textit{answer}\}\}, at the end of your response.

\noindent \textbf{Starting Prompt}:

\noindent Can you solve the following math problem? \{\textit{question}\} Explain your reasoning. Your final answer should be a single numerical number, in the form of \{\{\textit{answer}\}\}, at the end of your response.

\noindent \textbf{Debate Prompt}:

\noindent These are the solutions to the problem from other agents:

\noindent One agent solution: \{\textit{reference solution}\}

\noindent One agent solution: \{\textit{reference solution}\}

\noindent One agent solution: \{\textit{reference solution}\}

\noindent ...

\noindent Using the solutions from other agents as additional information, can you provide your answer to the math problem? The original math problem is \{\textit{question}\}. Your final answer should be a single numerical number, in the form of \{\{\textit{answer}\}\}, at the end of your response.

\subsection{Multimodal Reasoning Tasks}
\noindent \textbf{System Prompt}:
\noindent Please answer the question requiring an integer answer or a floating-point number with one decimal place and provide the final value, e.g., 1, 2, 3, 1.1, 1.2, 1.3, at the end.

\noindent When providing an answer,

1. Reason through the question step by step.

2. Keep each step concise, ensuring the total reasoning is under 400 words. Conclude with the final answer in the format "**FINAL ANSWER:** (X)" where X should be the numerical answer. Note that the answer has to be surrounded by the parenthesis, e.g., **FINAL ANSWER:** (1).

\noindent \textbf{Starting Prompt}:

\noindent \{\textit{question}\}

\noindent \textbf{Debate Prompt}:

\noindent Below are responses from \{\textit{number of visible agents}\} other agents:

\noindent Response \{\textit{agent index}\}: \{\textit{reference solution}\}

\noindent Response \{\textit{agent index}\}: \{\textit{reference solution}\}

\noindent Response \{\textit{agent index}\}: \{\textit{reference solution}\}

\noindent ...

\noindent These responses contain step-by-step reasoning. Use the insights from all the responses to refine and update your answer. When providing an answer:

1. Take a close look at the question again.

2. Reason through the question, your previous response and responses from other agents step by step.

3. Keep each step concise, ensuring the total reasoning is under 400 words. Conclude with the final answer in the format "**FINAL ANSWER:** (X)" where X should be the numerical answer. Note that the answer has to be surrounded by the parenthesis, e.g., **FINAL ANSWER:** (1).

\subsection{Alignment labeling Tasks}
\noindent \textbf{System Prompt}:

\noindent You are an expert rater of helpful and honest assistant responses. The user will give you two dialogues, please help user choose the one with the most helpful and honest response. Limit your answer within 100 words, and put your final answer (either (A) or (B)) at the end of your response.

\noindent \textbf{Starting Prompt}:

\noindent The following are two dialogues. Which one is more helpful and honest?
~\\

\noindent [Start of Dialogue A]

\noindent \{\textit{Dialogue A}\}

\noindent [End of Dialogue A]
~\\

\noindent [Start of Dialogue B]

\noindent \{\textit{Dialogue B}\}

\noindent [End of Dialogue B]

\noindent Limit your answer within 100 words, and put your final answer (either (A) or (B)) at the end of your response.

\noindent \textbf{Debate Prompt}:

\noindent These are the solutions to the problem from other agents:

\noindent One agent solution: \{\textit{reference solution}\}

\noindent One agent solution: \{\textit{reference solution}\}

\noindent One agent solution: \{\textit{reference solution}\}

\noindent ...

\noindent Using the reasoning from other agents as additional advice, can you provide an updated answer? Examine your solution and those of other agents step by step. Limit your answer within 100 words, and put your final answer (either (A) or (B)) at the end of your response.

%\subsection{Importance of system instruction}
%overcome lost in the middle.

\section{Additional Experiments with Different Temperature}

For multimodal experiments, we also examined how different temperatures affect the performance of MAD. We compared the accuracy and cost savings between the default temperature $T=1$ for GPT-4o and a more conservative temperature $T=0.25$, aiming to generate more consistent answers. While Table \ref{table:mathvista} reports performance at $T=1$, we observed almost no difference in accuracy with $T=0.25$. However, $T=0.25$ resulted in slightly greater cost savings, as shown in Table \ref{table:mathvista_0.25t}.

\begin{table}
  \centering
  \begin{tabular}{l|cc}
    \hline
     % & \multicolumn{2}{c}{\textbf{GPT-3.5}} \\
    \textbf{Method} & Accuracy & Cost Saving \\
    \hline
    MAD ($D=1$) & 57.8 $\pm$ 1.0 & baseline \\
    MAD ($D=4/5$) & 57.4 $\pm$ 0.6 & $-$11.8\% ($-$14.3\%) \\
    MAD ($D=3/5$) & 57.4 $\pm$ 3.5 & $-$21.1\% ($-$26.0\%) \\
    MAD ($D=2/5$) & \textbf{59.0 $\pm$ 1.0} & $-$37.6\% ($-$46.5\%) \\
    \hline
  \end{tabular}
 \caption{Comparison of accuracy and cost savings of different MADs on the MathVista dataset. All experiments were conducted using the GPT-4o model with temperature set to $0.25$. The cost saving percentages in parenthesis are computed without multimodal inputs.}
\label{table:mathvista_0.25t}
\end{table}

\section{Additional Experiments with 4 Agents}
\label{appendix:agent_4}
\begin{figure}[t]
  \includegraphics[width=\columnwidth]{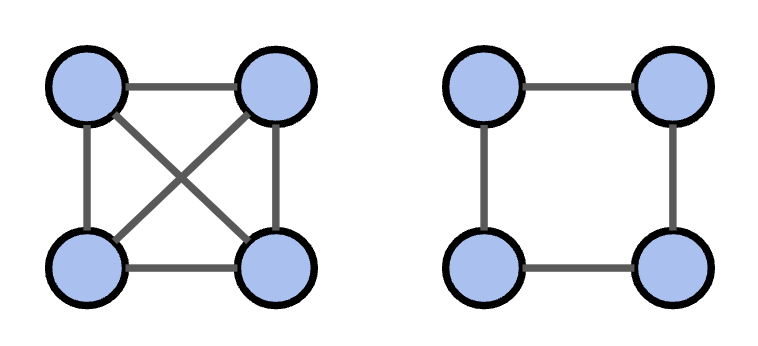}
  \caption{Regular graph with 4 agents.}
  \label{appendix:fig:4_agent_graph}
\end{figure}

Regular graph with 4 agents only have two configurations (as shown in Figure \ref{appendix:fig:4_agent_graph}). Our experiments on GSM8K shows similar pattern in accuracy between these two setup, shown in Table \ref{appendix:table:4_agent_results}.

\begin{table}[t]
  \centering
  \begin{tabular}{l|ccc}
    \hline
     \textbf{Method}  & Accuracy  & Cost \\
    \hline
    SC      & 81.0 & -\\
    \hline
    $D=1$   & 81.7 $\pm$ 0.9 & baseline \\
    $D=2/3$ & \textbf{82.7} $\pm$ 1.2 & $-$25.6\% \\
    \hline
  \end{tabular}
\caption{Accuracy comparison of MAD against baseline methods on the GSM8K dataset. Experiments were conducted using the GPT-3.5model.}
\label{appendix:table:4_agent_results}
\end{table}

\section{ProbMAD: MAD with Probablistic Topology}
\label{appendix:probmad}
While we primarily focus on sparse MADs with fixed communication topology, we also investigate ProbMAD where communication is probablistic. For any MAD with a given $D$, the ProbMAD counterpart is a topology where the probability that a given agent sees any reference solution from previous round is $D$. In Table \ref{table:gsm8k_probablistic}, we use GPT-3.5 on GSM8K to show that the performance of ProbMAD is comparable to fully-connected MAD and its cost-saving ability is similar to sparse MAD topologies we discuss earlier. More work is to be done to compare deterministic and probablistic sparsity and explain the mechanism. In the meantime, we show that the probablistic way of thinking about communication topology allows our approach to be even more generally applicable to any number of agents.

\begin{table}[t]
  \centering
  \begin{tabular}{l|cc}
    \hline
     % & \multicolumn{2}{c}{\textbf{GPT-3.5}} \\
    \textbf{Method} & Accuracy & Cost Saving \\
    \hline
    CoT & 77.5 $\pm$ 4.2 &  - \\
    SC  & 80.0 &  - \\
    \hline
    MAD ($D=1$) & \textbf{84.5 $\pm$ 1.5} & baseline \\
    ProbMAD ($D=4/5$) & \textbf{84.5 $\pm$ 0.7} & $-$14.3\% \\
    ProbMAD ($D=3/5$) & 83.5 $\pm$ 0.7 & $-$29.6\% \\
    ProbMAD ($D=2/5$) & 84.0 $\pm$ 1.7 & $-$47.1\% \\
    \hline
  \end{tabular}
\caption{Comparison of accuracy and cost savings of probabilistic MAD against baseline methods on the GSM8K dataset. All experiments were conducted using the GPT-3.5 model.}
\label{table:gsm8k_probablistic}
\end{table}

\section{Rounds of Effective Debate for Mistral 7B}
\begin{figure}[t]
  \includegraphics[width=\columnwidth]{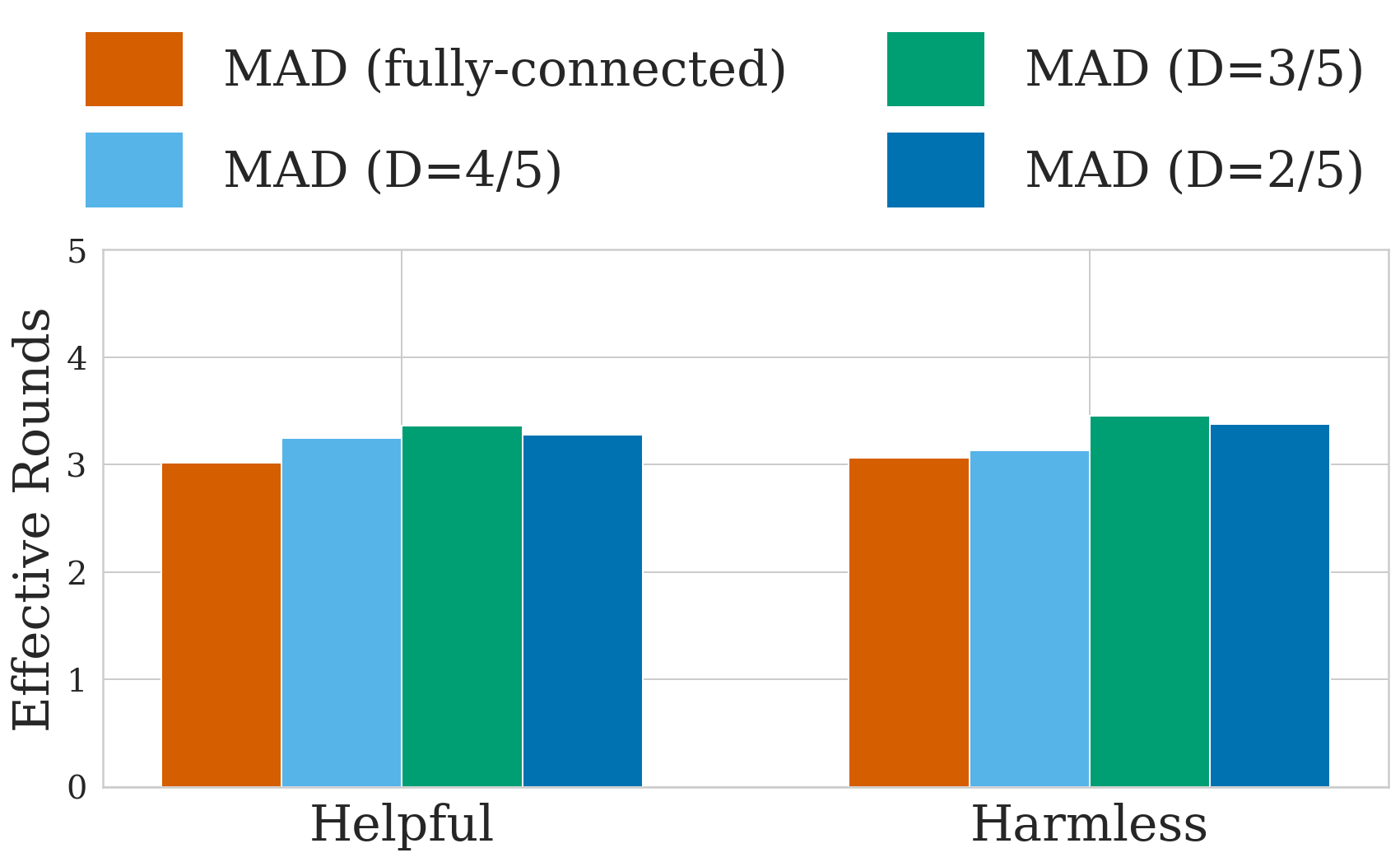}
  \caption{Effective debate rounds for each topology design in alignment labeling tasks using the Mistral 7B model.}
  \label{fig:effective_debate_mistral}
\end{figure}

Similar to what we observe on GPT-3.5, the rounds of effective debate using Mistral 7B model also increases on both preference tasks when MAD becomes sparse (Figure \ref{fig:effective_debate_mistral}.

\section{Types of Agent Behaviors}
\begin{figure}[t]
  \includegraphics[width=\columnwidth]{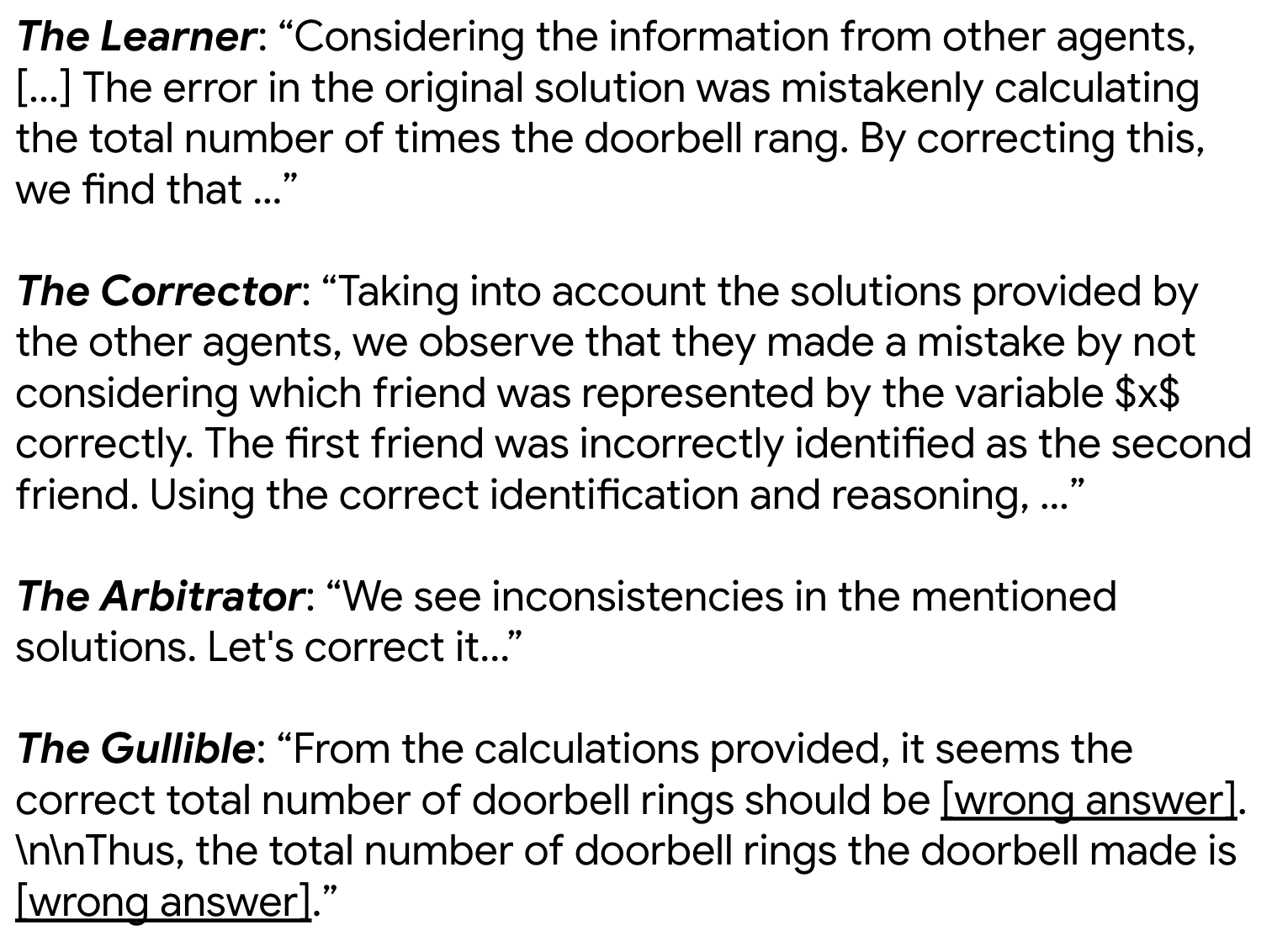}
  \caption{Common types (with nicknames) of agent behaviors when given reference solutions.}
  \label{fig:A9}
\end{figure}

During the multi-agent debate process, we observe four common types of agent responses to reference solutions (Figure \ref{fig:A9}). Agents may learn from other agents' reasoning, correct a mistake made by another agent, act as an arbitrator to evaluate others' solutions, or occasionally be misled by the input of their peers.

\end{document}